\documentclass[letterpaper]{article} 
\usepackage{aaai23}  
\usepackage{times}  
\usepackage{helvet}  
\usepackage{courier}  
\usepackage[hyphens]{url}  
\usepackage{graphicx} 
\usepackage{natbib}  
\usepackage{caption} 
\usepackage{algorithm}
\usepackage{algorithmic}

\usepackage{newfloat}
\usepackage{listings}
\DeclareCaptionStyle{ruled}{labelfont=normalfont,labelsep=colon,strut=off} 
\floatstyle{ruled}
\newfloat{listing}{tb}{lst}{}
\floatname{listing}{Listing}
\title{AdaptKeyBERT: An Attention-Based approach towards Few-Shot \& Zero-Shot Domain Adaptation of KeyBERT}
\author{
    Aman Priyanshu\textsuperscript{\rm 1}\equalcontrib,
    Supriti Vijay\textsuperscript{\rm 2}\equalcontrib,
}
\affiliations{
    \textsuperscript{\rm 1}Department of Information \& Communication Technology\\
    \textsuperscript{\rm 2}Department of Computer Science\\
    Manipal Institute of Technology,\\
    Manipal Academy of Higher Education,\\
    Karnataka, India.\\
    supriti.vijay@gmail.com
}

\usepackage{bibentry}

\begin{document}

\maketitle

\begin{abstract}
Keyword extraction has been an important topic for modern natural language processing. With its applications ranging from ontology generation, fact verification in summarized text, and recommendation systems. While it has had significant data-intensive applications, it is often hampered when the data set is small. Downstream training for keyword extractors is a lengthy process and requires a significant amount of data. Recently, Few-shot Learning (FSL) and Zero-Shot Learning (ZSL) have been proposed to tackle this problem. Therefore, we propose AdaptKeyBERT, a pipeline for training keyword extractors with LLM bases by incorporating the concept of regularized attention into a pre-training phase for downstream domain adaptation. As we believe our work has implications to be utilized in the pipeline of FSL/ZSL and keyword extraction, we open-source our code as well as provide the fine-tuning library of the same name AdaptKeyBERT at \url{https://github.com/AmanPriyanshu/AdaptKeyBERT}.
\end{abstract}

\section{Introduction}

The process of choosing the most important, pertinent, and descriptive terms from a single text to use as keywords is known as keyword extraction. In order to effectively transmit the informative content load of a document, keywords must represent distinct and specialized notions. Distinct keywords reduce redundancy in overall keywords extracted, whereas specialization assures quality keywords. However, with its varied applications in the fields of Bio-Medicine, Legal Entity Extractions, and News Verification among others, keyword extractors have become an integral segment of automated knowledge extraction systems. They've seen implementations for a varied resource of tasks such as summarization, indexing, contextual advertising, and personalized recommendations. Modern techniques have employed large-scale language model architectures, most having transformers as their backbone. Despite being quite effective at completing said task, they do need a considerable amount of labelled data for training, which may be difficult to source in particular low-resource domains and languages. Few-Shot Learning (FSL) \& Zero-Shot Learning (ZSL) offers an augmentable solution to the aforementioned problem.

\begin{figure}[t]
\centering
\includegraphics[width=0.9\columnwidth]{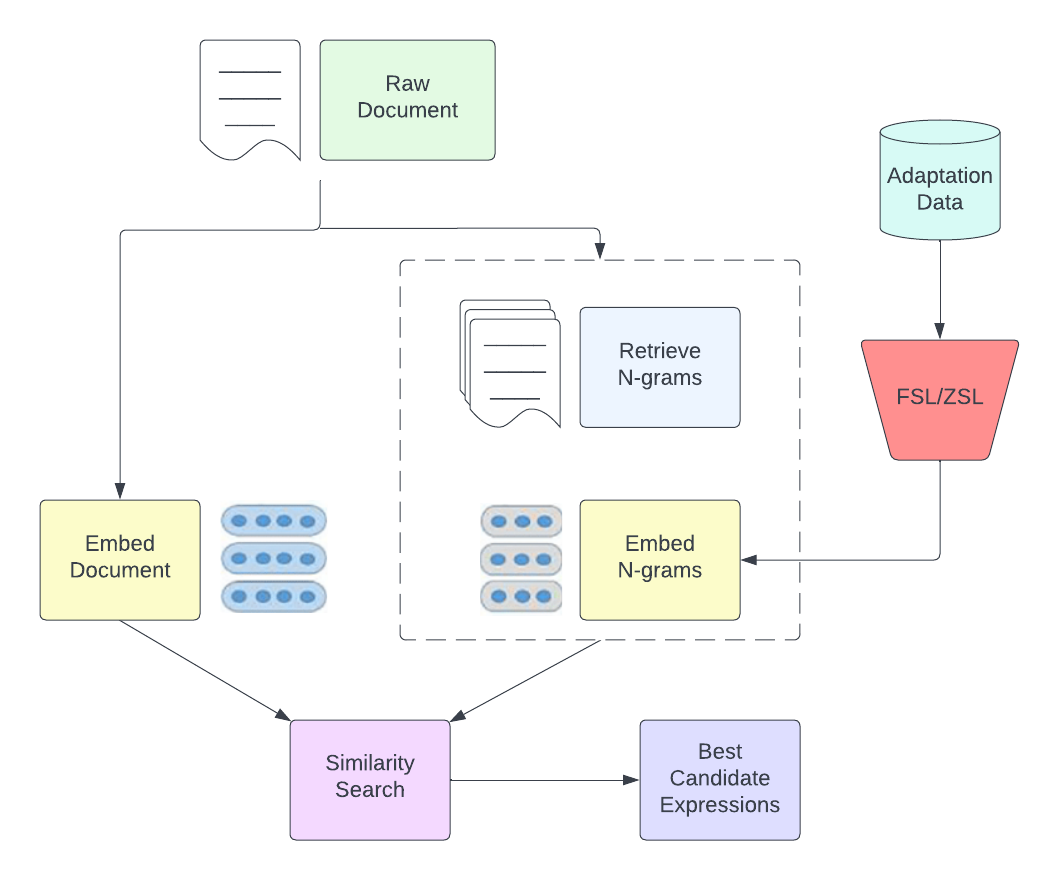}
\caption{Overall framework for zero-shot or few-shot integration to KeyBERT.}.
\label{fig1}
\end{figure}

To address this, we propose AdaptKeyBERT, which aims to integrate few-shot andd zero-shot learning through attention mapping over candidate embeddings. We present the overall progression of few-shot and zero-shot domain adaptation in Figure~\ref{fig1}. By incorporating regularized attention within KeyBERT's training module, we streamline the process of domain adaptation. We provide two variations to the algorithm, allowing users to encompass both (1) Few-Shot Domain Adaptation (2) Zero-Shot Domain Inclination.

An empirical study is then conducted to verify the nature of knowledge adapted after domain adaptation. Our objectives of achieving improved performance for both few-shot and zero-shot adaptation are formulated and described in detail below. Unlike existing frameworks for keyword extraction, AdaptKeyBERT exploits the significantly high performance of pre-trained LLM models present within KeyBERT, while improving them to qualitatively retrieve domain-specific keywords. Our contributions to this manuscript can be summarized as:

\begin{enumerate}
    \item We reconsider downstream training keyword extractors on varied domains by integrating pre-trained LLMs with Few-Shot and Zero-Shot paradigms for domain accommodation. We incorporate regularized attention based embedding reconstruction for domain attentive keyword extraction.
    \item Demonstrate two experimental setting with the objectives of achieving high performance for (a) Few-Shot Domain Adaptation, \& (b) Zero-Shot Domain Adaptation. The experimental results on both objective benchmarks demonstrate that our framework outperforms the base/naive approaches.
    \item We open source a python library (AdaptKeyBERT) for the construction of FSL/ZSL for keyword extraction models that employ LLMs directly integrated with the KeyBERT API. Allowing easy training, validation, and deployment of said models.
\end{enumerate}

\section{Related Work}
\subsection{Keyword Extraction}
Keyword extraction has become an indispensable tool for natural language processing. It has vast applications in downstream tasks like information retrieval systems - automatic text detection in historical manuscripts, slot filling, and extracting relations between entities, to name a few. 

Keyword extraction approaches vary from supervised to unsupervised methods, wherein the supervised techniques require a substantial amount of labelled data which proves to be expensive. On the other hand, unsupervised approaches are utilized in a low-resource scenario. Past work has focussed chiefly on unsupervised techniques, employing statistical (KPMiner \cite{ELBELTAGY2009132}, RAKE \cite{doi:https://doi.org/10.1002/9780470689646.ch1}, and YAKE \cite{CAMPOS2020257}), graph-based (TextRank \cite{mihalcea-tarau-2004-textrank}), embeddings-based (Key2Vec \cite{mahata-etal-2018-key2vec}, and EmbedRank \cite{doi:https://doi.org/10.1002/9780470689646.ch1}), KeyBERT \cite{grootendorst2020keybert}) and language-based methods \cite{tomokiyo-hurst-2003-language}. While supervised methods mainly delve into transformer architectures \cite{https://doi.org/10.48550/arxiv.1706.03762}.

Out of all the state-of-the-art methodologies, KeyBERT\cite{grootendorst2020keybert} has been widely accepted for its better performance. It leverages pretrained BERT-based embeddings for keyword extraction. These are ranked according to the cosine similarity and compared to the embedding of the entire document. Due to the better performance, we employ KeyBERT in our pipeline for keyword extraction.

\subsection{Keyword extraction with Zero-Shot Learning and Few Shot Learning}

Recent works have incorporated keyword extraction with zero-shot learning for document filtering, text detection and crosslinguistic( BiKEA \cite{huang-etal-2014-cross},  Takasu \cite{10.1145/1869446.1869454}). \cite{tigunova-etal-2020-charm} uses CHARM, a neural approach to identify salient keywords for a specific attribute. While \cite{ijcai2020p556}
proposes AGMHT, a zero-shot multi-label classifier that generates semantically meaningful features for zero-shot codes and reconstructs the code-relevant keywords. \cite{https://doi.org/10.48550/arxiv.2202.06650} employ zero-shot cross-lingual keyword detection and compare performance statistics amongst several state-of-the-art unsupervised keyword extractors. 

Keyword extraction with few-shot learning has been marginally explored. \cite{Gong_2021} detects new relations from unlabelled data, which utilizes weighted side information constructed from labels, keywords and hypernyms of entities. \cite{10.1145/3419972} introduces an attention-based deep relevance model for few-shot document filtering, ADRM. This outperforms DAZER \cite{li-etal-2018-deep}, a classifier with the same architecture but employed with a zero-shot setting.

In our pipeline, we employ both few-shot and zero-shot settings on our domain attentive keyword extractor to address the lack of labelled data in an unsupervised approach. To our knowledge, this is the first methodology that compares both settings (FSL \& ZSL) and incorporates them directly into keyword extraction architecture. Past work has only explored the integration of zero-shot learning in keyword extractors for cross-lingual datasets \cite{https://doi.org/10.48550/arxiv.2202.06650}.

\section{Methodology}
\label{adaptkeybertmethod}

\subsection{Problem Formulation}

As illustrated in Figure~\ref{fig1}, AdaptKeyBERT encapsulates zero-shot and few-shot learning by recomputing the embedding vectors of the candidate words. We first detail the notations and functioning of the baseline KeyBERT algorithm. For a keyword extraction task, we consider a target document $\textbf{D}$ from which we aim to extract relevant keywords. $\textbf{D}$ consists of multiple words/phrases, each of which could be potentially relevant keyword. Therefore, KeyBERT preliminarily employs n-gram based feature extraction. Here, we define the range of the n-gram extractor as, $[n_{ini}, n_{fin}]$. It defines the lower and upper boundary of the range of n-values for different word n-grams or char n-grams to be extracted. Once extracted we consider these as samples for the candidate set, $\textbf{C} = \{c_1, c_2, c_3, ..., c_k\}$. Therefore, the task now encapsulates into ranking the above candidates in order of relevancy, such that they're still distinct from one-another. To solve the \textit{ranking} function KeyBERT employs a Sentence Embedding network, $\textbf{EModel}$, which encodes sentences/phrases into high dimensional vectors. This in-turn is used to compute the latent vector representation of the document, $\textbf{D} \rightarrow E_D$, and the candidate set, $\textbf{C} \rightarrow E_{C}$. Where, $E_{C} = \{e_{c_1}, e_{c_2}, e_{c_3}, ..., e_{c_k}\}$ computed for each candidate phrase present. We then aim to maximize the ability of the ranking system based on their similarity to the document embedding.

\subsubsection{Few-Shot Integration}

Given a keyword extraction task, it becomes integral to consider domain adaptation as an integral aspect of the field. With the generalization of Sentence Embedding Networks over all fields, it also reflects poorly on disproportionately marginalized topics. For example, we do not expect certain topics, such as core biology taxonomy, to be present in normal training sets or words which are relevant to legal artefacts. In such cases, the ability to rank certain domain words higher would be highly beneficial. Therefore, a methodology to encompass the above objective is required, with same a smaller sample space.

Given the above KeyBERT module with an objective of few-shot domain adaptation, we employ attention to re-compute candidate-set embeddings. We first create a pre-training step, where the user must feed the model with sample $(\textbf{D}, \textbf{RC})$ tuples. Where each document, $\textbf{D}$ is paired with its own list of relevant candidate phrases, $\textbf{RC} \rightarrow \{rc_1, rc_2, rc_3, ... rc_l\}$. The embedding extraction mechanism of KeyBERT remains the same during this phase, however, we append a trainable \textit{attention-layer} to the model, which aims to optimize the following function:

\begin{equation}
    L(\theta)_{c \in RC} = \sum_{i=1}^{l} MSE(E_D, a_{c_i})
\end{equation}

Where the model learns to adapt the attention-computed embedding vector ($a_c$) of relevant keywords to be closer to the document embeddings, \textbf{D}, which in-turn are domain relevant. This allows the model to adapt to the topical shit. At the same time, we do not wish dissociate the embeddings of the other keywords, as they may be relevant in their own right. At the same time, restricting them to their original latent representation, allows AdaptKeyBERT to ensure that the attention doesn't simply move all the computed vectors towards the document embedding without semantic context. For this we employ the loss, 

\begin{equation}
    L(\theta)_{c \notin RC} = \sum_{i=1}^{l} MSE(a_{c_i}, e_{c_i})
\end{equation}

This allows the attention mechanism to re-compute the embedding vector such that relevant keywords $E_{RC}$ have an latent representation closer to the document embedding $E_D$, thus giving them an advantage over other candidate sets, which remain the same.

\subsubsection{Zero-Shot Integration}

With the construction of a few-shot methodology for domain-adaptation, we observe that certain words can simply account as seed for words for particular domains, $\textbf{SC} \rightarrow \{sc_1, sc_2, sc_3, ..., sc_m\}$. However, the methodology currently presented in KeyBERT among other keyword extraction techniques employing LLMs employ weighted mean computation between the original document embedding and the seed word embeddings ($E_{SC}$). Therefore, AdaptKeyBERT employs attention-computed weights to ascertain the relevance of candidate words over the seed words. A maximal score is then taken and used as the \textit{similarity-weight}, $sw$, for re-computing the embeddings, $a_c$. A regularizer, $\alpha \in [0, 1]$, for reconfiguring the significance of the zero-shot integration methodology is also defined. We define this by, 

\begin{equation}
    a_c = (1 - sw * \alpha) * e_c + (sw * \alpha) * E_D
\end{equation}

\subsection{Implementation Details}

We build on both the objectives mentioned above, making sure AdaptKeyBERT delivers for both few-shot learning as well as zero-shot learning. For a given task, we ascertain the most popular words presented in the dataset. These are directly fed into the zero-shot module as seed words for its integration. On the other hand, we retrieve those samples which encompass most of the these popular keywords within themselves from the dataset for few-shot learning. A popular keyword here is defined as being present in more than $p\%$ of the dataset (here, $p\%=10\%$). For few-shot learning we restrict number of samples fed for domain adaptation to $p\%$ as well. Our experiments demonstrate the ability of the model to generalize on new toics only from a subset of the original data.

\section{Experiments}

\subsection{Datasets and Experimental Settings}

We categorize our experiments into two benchmarks based on the aforementioned objectives. We discuss the \textbf{fao780 dataset} (Food and Agriculture Organization). Which consists of 780 documents from the food and agriculture industry. We also take a look at the \textbf{CERN-290 dataset} which is composed of 290 high energy physics documents. We first quantify the most popular words in each dataset, allowing is to leverage AdaptKeyBERT to better understand domain adaptation for these particular domains.

\subsection{Results}

\begin{table}[h!]
\begin{tabular}{|l|l|l|l|}
\hline
\textbf{Model} & \textbf{Precision} & \textbf{Recall} & \textbf{F-Score} \\ \hline
\textbf{Benchmark} & 36.74 & 33.67 & 35.138 \\ \hline
\textbf{Zero-Shot} & 37.25 & 38.59 & 37.908 \\ \hline
\textbf{Few-Shot} & 40.03 & 39.1 & 39.559 \\ \hline
\textbf{\begin{tabular}[c]{@{}l@{}}Zero-Shot \\ \& Few-Shot\end{tabular}} & 40.02 & 39.86 & 39.938 \\ \hline
\end{tabular}
\caption{AdaptKeyBERT performance on FAO-780 dataset with $p\%=10\%$.}
\label{tab:2}
\end{table}

We perform a progressive learning routine to compute the performance of AdaptKeyBERT against the preliminary approach of KeyBERT. For the formulation of naive approach, we assume SentenceEmbeddings from the $all-MiniLM-L6-v2$ as the baseline model. We benchmark the performance of this model for comparison. We present results over both the datasets, FAO-780 dataset and the CERN-290 dataset.

We realize congruent results for both the datasets. Focusing on the FAO-780 results presented in Table~\ref{tab:2} we can see a distinguishing improvement. While the Zero-Shot model weakly outperforms its benchmark counterpart compared to the Few-Shot approach. There's an improvement of 7.88\% and 12.58\% F1-Score respectively. On the contrary however, the combination of the two methodologies fails to improve drastically over the performance of the Few-Shot module. This may have been caused due to unintentional bias that similarity computation might have enabled contributed by the Zero-Shot framework. Even so, the amalgamation of both techniques achieves an improvement of over 13.66\%.

\begin{table}[h!]
\begin{tabular}{|l|l|l|l|}
\hline
\textbf{Model} & \textbf{Precision} & \textbf{Recall} & \textbf{F-Score} \\ \hline
\textbf{Benchmark} & 24.74 & 26.58 & 25.627 \\ \hline
\textbf{Zero-Shot} & 27.35 & 25.9 & 26.605 \\ \hline
\textbf{Few-Shot} & 29 & 27.4 & 28.177 \\ \hline
\textbf{\begin{tabular}[c]{@{}l@{}}Zero-Shot \\ \& Few-Shot\end{tabular}} & 29.11 & 28.67 & 28.883 \\ \hline
\end{tabular}
\caption{AdaptKeyBERT performance on CERN-290 dataset with $p\%=10\%$.}
\label{tab:1}
\end{table}

Taking into consideration the CERN-290 dataset, we discover drastic improvements over the precision metric. However, the same fails for the recall as the number of False-Negatives increases in this benchmark. Even so, we achieve considerable performance for the same, improving F1-Score by 3.81\%, 9.95\%, and 12.7\% respectively for Zero-Shot, Few-Shot, Zero+Few-Shot integration.

\section{Conclusion}

With an objective to adapt over new \& disproportionately represented domains, we introduce AdaptKeyBERT, a first of its kind open-sourced pipeline which extends the KeyBERT library. By providing results across two liguistic-heavy datasets focused on different domains, we validate the performance and adaptation ability of our proposed methodology. Since keyword extraction models are often the first step in the construction of knowledge bases as well as the identification of certain information, we believe AdaptKeyBERT will be able to save a significant amount of memory and time computationally. We acknowledge that our keyword extraction pipeline may be susceptible to traditional biases plaguing similar ideations \cite{firoozeh_nazarenko_alizon_daille_2020}, which we aim to address in future work.

\bibliography{aaai23}

\end{document}